\documentclass[runningheads]{llncs}
\usepackage[T1]{fontenc}
\usepackage{graphicx,verbatim}
\usepackage{epsfig}
\usepackage{graphicx}
\usepackage{amsmath}
\usepackage{amssymb}
\usepackage{booktabs}
\usepackage{multirow}
\usepackage{hyperref}
\usepackage{xcolor}

\usepackage{array} 
\usepackage{graphicx} 
\usepackage{tabularx}
\usepackage{ragged2e}
\usepackage[table]{xcolor}
\usepackage{subcaption}
\usepackage{tikz}
\usetikzlibrary{calc}
\usepackage{xcolor}
\usepackage{tabularx,booktabs,array,xcolor}
\usepackage{ragged2e}
\usepackage{pifont}
\usepackage{marvosym} 

\definecolor{good}{RGB}{34,139,34}  
\definecolor{bad}{RGB}{200,40,40}   
\definecolor{badgebg}{RGB}{250,250,250}
\newcolumntype{L}[1]{>{\raggedright\arraybackslash}p{#1}}
\newcolumntype{C}[1]{>{\centering\arraybackslash}p{#1}}

\begin{document}
\title{MedObvious: Exposing the Medical Moravec's Paradox in VLMs via Clinical Triage}
\titlerunning{MedObvious}
%
\author{Ufaq Khan\thanks{Corresponding author}$^{1}$
\and Umair Nawaz$^{1}$
\and Lekkala Sai Teja$^{2}$
\and Numaan Saeed$^{1}$
\and Muhammad Bilal$^{3}$
\and Yutong Xie$^{1}$
\and Mohammad Yaqub$^{1}$
\and Muhammad Haris Khan$^{1}$}

\authorrunning{Khan et al.}  

\institute{
$^{1}$ Mohamed bin Zayed University of Artificial Intelligence, UAE \\
$^{2}$ National Institute of Technology, Silchar \\
$^{3}$ Birmingham City University, UK \\
\Letter\ \email{ufaq.khan@mbzuai.ac.ae} \\
\vspace{0.4em}
Project Page: \url{https://ufaqkhan.github.io/MedObvious-Website/}
}

\maketitle
\begin{abstract}
Vision Language Models (VLMs) are increasingly used for tasks like medical report generation and visual question answering. However, fluent diagnostic text does not guarantee safe visual understanding. In clinical practice, interpretation begins with \emph{pre-diagnostic sanity checks}: verifying that the input is valid to read (correct modality and anatomy, plausible viewpoint and orientation, and no obvious integrity violations). Existing benchmarks largely assume this step is solved, and therefore miss a critical failure mode: a model can produce plausible narratives even when the input is inconsistent or invalid. We introduce \textbf{MedObvious}, a 1,880-task benchmark that isolates input validation as a set-level consistency capability over small multi-panel image sets: the model must identify whether any panel violates expected coherence. MedObvious spans five progressive tiers, from basic orientation/modality mismatches to clinically motivated anatomy/viewpoint verification and triage-style cues, and includes five evaluation formats to test robustness across interfaces. Evaluating 17 different VLMs, we find that sanity checking remains unreliable: several models hallucinate anomalies on normal (negative-control) inputs, performance degrades when scaling to larger image sets, and measured accuracy varies substantially between multiple-choice and open-ended settings. These results show that pre-diagnostic verification remains unsolved for medical VLMs and should be treated as a distinct, safety-critical capability before deployment.


\end{abstract}

\begin{figure*}[t]
\centering
\begin{subfigure}[t]{0.32\textwidth}
  \centering
  \caption*{(a) Visual Referring}
  \includegraphics[width=0.8\linewidth]{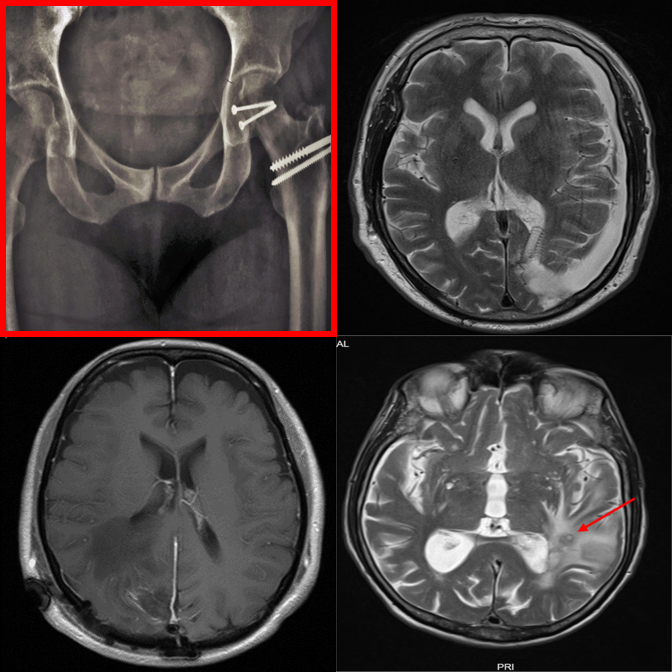}
  \label{fig:qual_a}
\end{subfigure}
\begin{subfigure}[t]{0.32\textwidth}
  \centering
  \caption*{(b) Detection MCQ}
  \includegraphics[width=0.8\linewidth]{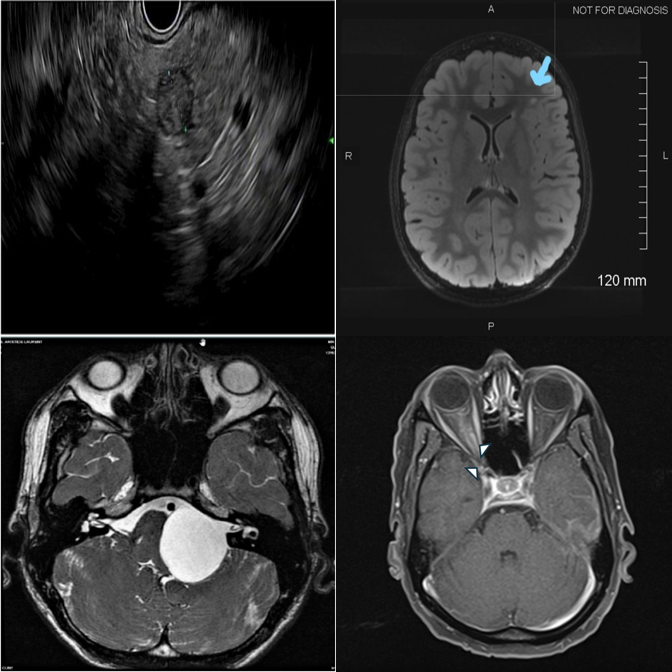}
  \label{fig:qual_b}
\end{subfigure}
\begin{subfigure}[t]{0.32\textwidth}
  \centering
  \caption*{(c) Detection MCQ}
  \includegraphics[width=0.8\linewidth]{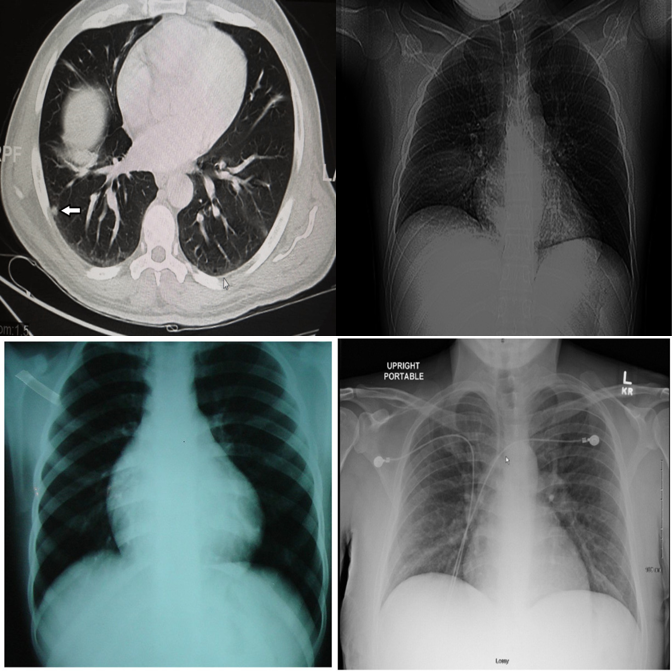}
  \label{fig:qual_c}
\end{subfigure}

\scriptsize
\setlength{\tabcolsep}{4.5pt}
\renewcommand{\arraystretch}{1.15}
\rowcolors{2}{black!3}{white}
\newcolumntype{L}[1]{>{\raggedright\arraybackslash}p{#1}}
\newcolumntype{C}[1]{>{\centering\arraybackslash}p{#1}}
\resizebox{\textwidth}{!}{%
\begin{tabular}{@{} L{0.15\textwidth} C{0.3\textwidth} C{0.3\textwidth} C{0.3\textwidth} @{}}
\toprule
 & \textbf{(a)} & \textbf{(b)} & \textbf{(c)} \\
\midrule
\textbf{Question} &
Is this highlighted scan the clinical outlier; i.e., does it differ in modality, anatomy, or pathology from the others? Answer yes/no &
You are reviewing a $2\times2$ grid of medical scans. One scan is the clinical outlier as it differs in modality, anatomy, or pathology. Which position contains the outlier?
\textit{A)} top-left \textit{B)} top-right \textit{C)} bottom-left \textit{D)} bottom-right \textit{E)} None &
In this $2\times2$ medical image grid, one scan contains a clinical anomaly (chest CT). Which position is it?
\textit{A)} top-left \textit{B)} top-right \textit{C)} bottom-left \textit{D)} bottom-right \textit{E)} None \\
\textbf{Ground truth} & Yes & A (Top-left) & A (Top-left) \\
\textbf{LLaVA-1.5} & No \textcolor{red!70!black}{(\ding{55})} & B \textcolor{red!70!black}{(\ding{55})} & B \textcolor{red!70!black}{(\ding{55})} \\
\textbf{Qwen3-VL} & Yes \textcolor{green!55!black}{(\ding{51})} & D \textcolor{red!70!black}{(\ding{55})} & D \textcolor{red!70!black}{(\ding{55})} \\
\textbf{Lingshu} & Yes \textcolor{green!55!black}{(\ding{51})} & B \textcolor{red!70!black}{(\ding{55})} & A \textcolor{green!55!black}{(\ding{51})} \\
\bottomrule
\end{tabular}
}
\caption{\textbf{Qualitative MedObvious examples for pre-diagnostic visual triage.}
Each column corresponds to a grid in (a)--(c). We report the task question, ground truth, and predictions from representative VLMs.}
\label{fig:main_benchmark_figure}
\end{figure*}

\section{Introduction}

Vision-Language Models (VLMs) are increasingly being used to interpret medical images. Recent systems can generate radiology-style descriptions, answer clinical questions, and perform multi-step reasoning over images and text, driven by both general-purpose models such as GPT-4o \cite{achiam2023gpt}, Flamingo~\cite{alayrac2022flamingo} and LLaVA~\cite{liu2023visual} and medical adaptations such as LLaVA-Med~\cite{li2023llava}, RadFM~\cite{wu2025towards}, and others \cite{pan2025medvlm,jiang2025omniv,nath2025vila}. In parallel, these models are being explored as the core perception for \textit{visual AI agents} \cite{lin2025showui,fallahpour2025medrax} that can also interact with imaging software (e.g., navigating viewers, selecting series, adjusting visualization, and triggering downstream tools). This progress has motivated their potential use as assistants for clinical reporting and decision support. However, fluent language generation does not guarantee reliable visual perception. VLMs may produce coherent diagnostic narratives while failing basic sanity checks, such as detecting incorrect orientation, mismatched anatomy, unexpected modality, or physically implausible artifacts. We refer to this mismatch as the \textbf{Medical Moravec's Paradox}, extending Moravec's observation~\cite{agrawal2010study} that perception and spatial reasoning, trivial for humans, can be disproportionately difficult for machines even when higher-level outputs appear plausible. In medical imaging, this gap is consequential because failures occur before diagnosis: when the input is invalid or inconsistent, downstream reports become clinically uninterpretable.

Clinical interpretation begins with pre-diagnostic triage: clinicians first verify body part, view, modality, laterality, orientation, and basic image integrity, and they do not proceed to diagnosis if these checks fail. This requirement is amplified in \textit{multi-image} and \textit{AI-agentic} settings, where decisions depend on consistency across a set of inputs, such as multiple fetal ultrasound views or long CT/MRI slice sequences and series. A single mis-acquired view, corrupted slice, mismatched series, or orientation error can compromise study-level reasoning, especially for models that aggregate evidence across images. 
Moreover, common tools such as 3D Slicer \cite{fedorov20123d} and ITK-SNAP \cite{yushkevich2006user} support multi-panel layouts (e.g., axial, sagittal, coronal, and 3D) similar to a $2\times2$ display. VLM-based agents operating in these viewers must detect obvious panel-level inconsistencies to avoid acting on the wrong series, anatomy, or invalid inputs. Despite their importance, pre-diagnostic competencies are rarely evaluated explicitly. Standard medical VLM benchmarks such as VQA-RAD~\cite{lau2018dataset}, PathVQA~\cite{he2020pathvqa}, PMC-VQA~\cite{zhang2023pmc}, VQA-Med~\cite{ben2019vqa}, and SLAKE~\cite{liu2021slake} primarily assess the correctness of final answers, while hallucination-focused benchmarks such as Med-HallMark~\cite{chen2024detecting} emphasize factual consistency of the generated text\cite{khan2024ultraweak}. These settings largely assume the input has been correctly perceived and can therefore miss failures on visually obvious sanity checks, allowing models to score well while remaining brittle and potentially unsafe in real multi-image or agentic workflows.

\begin{figure}[t]
    \centering
    \includegraphics[width=0.95\textwidth]{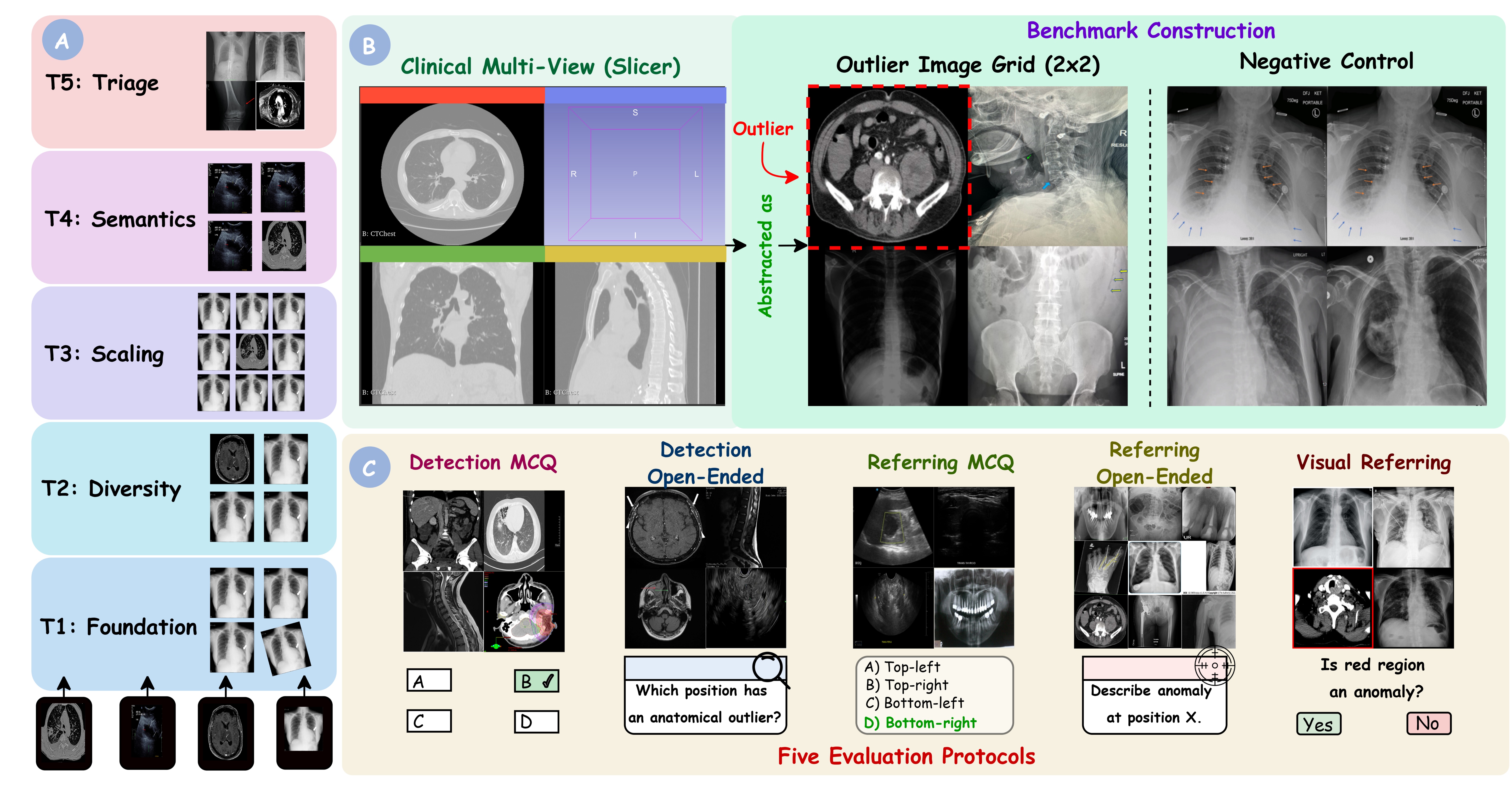}
    \caption{\textbf{MedObvious overview.} 
    (\textbf{A}) Five progressive tiers (T1--T5). 
    (\textbf{B}) Construction: multi-view studies are abstracted into small grids with either a single outlier or a negative-control (no outlier). 
    (\textbf{C}) Five evaluation protocols: Detection (MCQ/Open), Referring (MCQ/Open), and Visual Referring (Yes/No).}
    \label{fig:medobvious}
\end{figure}

We introduce \textbf{MedObvious}, a benchmark for pre-diagnostic visual sanity checking in medical images. MedObvious asks a question that precedes diagnosis: \textit{Is the input coherent and appropriate to interpret?}
For this, we present small grids ($2\times2$ or $3\times3$) and require models to either identify an outlier panel or state that no outlier exists, as shown in Fig.~\ref{fig:main_benchmark_figure}. Although it is not a clinical reading interface, this provides a controlled probe of set-level consistency, mirroring requirements in multi-view ultrasound, multi-slice CT/MRI, and multi-panel viewer agentic workflows. MedObvious evaluates two axes. The \textit{Clinical Safety} axis targets real failure modes that should be caught before any diagnostic verdict, including wrong body parts, flipped images, viewpoint/anatomy mismatches, and grossly apparent major pathology. 
The \textit{Visual Grounding} axis uses synthetic inconsistencies (e.g., inserting modality-incompatible textures into an image) to test whether models check the visual input itself rather than relying on language priors for downstream tasks. MedObvious also includes explicit \textbf{negative controls} where all panels are consistent, so the correct response is that no outlier exists, directly measuring false alarms. Furthermore, it is organized into five progressive tiers (T1--T5) that increase in difficulty and clinical specificity as depicted in Fig.~\ref{fig:medobvious}, ranging from basic orientation/modality mismatches to anatomy/viewpoint inconsistencies and high-saliency triage failures.
In our zero-shot evaluation across \textit{7 general}, \textit{4 medical}, and \textit{6 proprietary} VLMs, performance remains uneven as the best mean accuracy reaches 63.2\%, yet negative-control accuracy spans a wide range,
indicating that false alarms on normal inputs remain common. We also observe strong format sensitivity, with large gaps between multiple-choice and open-ended variants of the same underlying capability. Our main contributions are:

\begin{itemize}
    \item We formalize the \textit{Medical Moravec's Paradox} for medical VLMs, highlighting a gap between fluent diagnostic language and reliable pre-diagnostic visual sanity-checking, especially in multi-image and agentic-viewer settings.
    \item We present first \textbf{MedObvious}, a 1{,}880 task benchmark spanning 5 progressive tiers, multiple grid configurations, five evaluation modes, and systematic negative controls, designed to evaluate pre-diagnostic visual triage independently of diagnosis.
    \item We benchmark 7 representative open-source, 6 closed-source, and 4 medically specialized VLMs under zero-shot inference. 
\end{itemize}


\section{MedObvious Construction}

MedObvious is a benchmark designed to test whether medical VLMs can recognize \emph{obvious} input-level inconsistencies before attempting any diagnosis, as inspired from \cite{dahou2025vision}. It acts as a prerequisite for safe medical VLM deployment by \emph{pre-diagnostic visual sanity checking}. Before interpreting pathology, clinicians first verify that an image is valid to read, including modality, anatomical region, viewpoint/orientation, and basic integrity. If these checks fail, the subsequent diagnosis is unreliable regardless of how fluent the generated text appears. MedObvious isolates this input-validation step and explicitly evaluates it.

\noindent \textbf{Motivation.} The clinical need is inherently \emph{set-based}. Many studies are interpreted across multiple views, slices, or series, where a single inconsistent element can compromise the conclusions of the study. In fetal ultrasound, assessment of the fetal heart relies on multiple standard views. One mis-acquired plane or corrupted view may change the interpretation. In CT/MRI, clinicians reason over long slice sequences and multi-series studies. In longitudinal assessment, such as reviewing the progression of Multiple Sclerosis from two or more brain MRI scans, an incorrectly positioned slice could lead to a completely different clinical decision. A flipped series, corrupted slice, or modality/anatomy mismatch is a coherence break that should be detected prior to diagnosis. This set-level checking is also reflected in common imaging software (e.g., 3D Slicer and ITK-SNAP), which presents multi-planar views in multi-panel layouts resembling $2\times2$ grid. Our grid-based tasks are therefore a controlled abstraction of this real requirement by detecting whether any element in a small visual set violates expected consistency before proceeding to downstream reasoning or agentic actions.
\begin{table}[hbtp]
\centering
\small
\setlength{\tabcolsep}{10pt}        
\renewcommand{\arraystretch}{1.08} 
\caption{\textbf{MedObvious composition.} Tiers increase in clinical specificity. 
}
\label{tab:benchmark_stats}
\resizebox{0.9\textwidth}{!}{%
\begin{tabular}{l l c c r r r p{4.8cm}}
\toprule
\rowcolor{blue!6}
\textbf{Tier} & \textbf{Name} & \textbf{Grid} & \textbf{\#Mod.} & \textbf{Tasks} & \textbf{Pos.} & \textbf{Neg.} & \textbf{Clinical analog} \\
\midrule
T1 & \textsc{Foundation} & $2\times2$ & 2 & 440 & 275 & 165 & Basic QC (orientation, modality) \\
T2 & \textsc{Diversity}  & $2\times2$ & 2 & 480 & 300 & 180 & Robustness across sources \\
T3 & \textsc{Scaling}    & $3\times3$ & 4 & 360 & 225 & 135 & Multi-view/-slice comparison \\
T4 & \textsc{Semantics}  & $2\times2$ & 3 & 320 & 200 & 120 & Anatomy/viewpoint verification \\
T5 & \textsc{Triage}     & Mixed      & 4 & 280 & 175 & 105 & ``Stop and verify'' safety cues \\
\midrule
\rowcolor{green!6} 
\textbf{Total} &  &  & \textbf{4} & \textbf{1{,}880} & \textbf{1{,}175} & \textbf{705} &  \\
\bottomrule
\end{tabular}
}
\end{table}

\noindent\textbf{Problem Formulation.} Each MedObvious instance presents a grid of $n$ images, $G=\{I_1,\ldots,I_n\}$ with $n\in\{4,9\}$. The model must either identify the index of the inconsistent image or state that no outlier exists ($y \in \{1,\ldots,n\}\cup\{\varnothing\}$)
where $y=k$ denotes $I_k$ as the outlier and $y=\varnothing$ denotes a clean and consistent grid. 
This setup evaluates input validity and set-level coherence, i.e., whether images that should belong to the same study context (across views, slices/series, or timepoints) are mutually consistent, rather than evaluating diagnosis.

\noindent \textbf{Datasets.} We primarily use curated subsets of ROCO~\cite{roco} to construct anatomically and modality-defined tasks (e.g., chest radiographs, CT, MRI, ultrasound) using metadata filtering. To reduce modality shortcuts and enforce modality awareness, we additionally include non-radiological images, including endoscopy from Kvasir \cite{pogorelov2017kvasir}, to increase visual diversity and discourage reliance on generic grayscale radiology priors.

\noindent \textbf{Template-based generation.}
Each task is generated from a shared template: we sample $n-1$ \emph{inlier} panels from a reference category (defined by modality, and when available, anatomy/viewpoint), and then either (i) insert one \emph{outlier} from a different category or via a controlled integrity violation (e.g., orientation change or physically inconsistent composite), or (ii) create a \emph{negative control} by sampling all $n$ panels from same reference category so the correct answer is $\varnothing$. 

\noindent\textbf{Progressive tiers.}
MedObvious comprises five tiers (1,880 tasks) with increasing clinical specificity (Table~\ref{tab:benchmark_stats}): T1 \textsc{Foundation} ($2\times2$; orientation/modality mismatches), T2 \textsc{Diversity} ($2\times2$; finer distinctions across sources), T3 \textsc{Scaling} ($3\times3$; many distractors and modality diversity), T4 \textsc{Semantics} ($2\times2$; anatomy/viewpoint mismatches and integrity violations), and T5 \textsc{Triage}; high-saliency failures (like gross abnormalities), and cross-modality coherence breaks.

\noindent \textbf{Evaluation Protocols.} Each grid is evaluated in 5 formats to separate visual capability from response-format effects: Detection MCQ (\textit{pick the outlier}), Detection Open (\textit{state the outlier position}), Referring MCQ (\textit{choose an outlier description given its position}), Referring Open (\textit{describe the outlier given its position}), and Visual Referring (\textit{Yes/No for highlighted region}). Using both multiple-choice and open-ended settings exposes selection biases and over-generation.

\noindent\textbf{Negative controls.} To measure false alarms, MedObvious includes explicit negative controls where all panels are consistent and no outlier exists (37.5\% of tasks; 705/1{,}880). The correct label is $y=\varnothing$ (or ``No'' for binary verification).

\begin{table*}[t]
\centering
\small
\caption{\textbf{MedObvious: Per-task accuracy (\%).} \textbf{Task A--E} correspond to Detection MCQ, Detection Open, Referring MCQ, Referring Open, and Visual Referring, respectively. ``Positive (Pos)''/``Negative (Neg)'' indicate accuracy on anomalous/non-anomalous samples. Best results per task are \textbf{bolded}.}
\label{tab:main_results}

\setlength{\tabcolsep}{7pt}
\renewcommand{\arraystretch}{1.08}
\resizebox{0.8\textwidth}{!}{%
\begin{tabular}{lccccc|cc|c}
\toprule
\textbf{Model} & \textbf{Task-A} & \textbf{Task-B} & \textbf{Task-C} & \textbf{Task-D} & \textbf{Task-E} & \textbf{Pos($+$)} & \textbf{Neg($-$)} & \textbf{Avg} \\
\midrule
\multicolumn{9}{c}{\textit{\cellcolor{cyan!15}General open-source VLMs}} \\
\midrule
LLaVA-1.5-7B    & 40.4 & 22.3 & 22.1 & 35.7 & 50.0 & 37.5 & 31.9 & 34.3\\
Qwen2-VL-7B     & 32.3 & 49.5 & 72.7 & 25.1 & 53.1 & 56.3 & 28.7 & 45.4\\
Qwen2.5-VL-7B   & \textbf{58.7} & \textbf{82.1} & 75.3 & 29.7 & \textbf{65.1} & 60.9 & \textbf{70.7} & \textbf{63.2}\\
Qwen3-VL-8B     & 31.2 & 32.7 & \textbf{80.8} & \textbf{38.7} & 52.9 & \textbf{68.9} & 2.9  & 44.0\\
InternVL2.5-8B  & 56.3 & 56.3 & 69.7 & 26.8 & 51.7 & 59.7 & 42.5 & 51.9\\
InternVL3-8B    & 38.5 & 43.6 & 80.4 & 27.6 & 50.2 & 64.6 & 16.6 & 45.9\\
Pixtral-12B     & 31.0 & 22.5 & 76.6 & 26.3 & 51.9 & 59.5 & 5.3  & 39.0\\
\midrule
\multicolumn{9}{c}{\textit{\cellcolor{green!15}Medical open-source VLMs}} \\
\midrule
LLaVA-Med-7B        & 10.0 & 36.8 & 21.2 & 23.4 & 50.0 & 37.1 & 17.5 & 28.0\\
Fleming-8B          & 26.8 & 23.8 & 78.3 & 23.8 & 50.0 & 57.4 & 5.3  & 37.9\\
Medgemma1.5-4B-IT   & 23.6 & \textbf{86.1} & 43.4 & 19.5 & \textbf{66.6} & 44.6 & \textbf{64.1} & 49.7\\
Lingshu-7B          & \textbf{39.3} & 78.5 & \textbf{79.5} & \textbf{26.8} & 61.9 & \textbf{66.8} & 43.8 & \textbf{56.6}\\
\midrule
\multicolumn{9}{c}{\textit{\cellcolor{orange!15}Proprietary VLMs}} \\
\midrule
Gemini-2.0-Flash & 54.2 & 42.7 & \textbf{85.9} & \textbf{35.7} & \textbf{69.3} & \textbf{75.4} & 25.6 & \textbf{55.5} \\
Gemini-2.5-Flash & \textbf{67.2} & 45.5 & 80.4 & 31.9 & 55.9 & 74.1 & \textbf{26.3} & 54.4 \\
GPT-4o           & 47.2 & \textbf{50.4} & 62.8 & 26.3 & 61.7 & 68.0 & 22.7 & 48.4 \\
GPT-4.1-nano     & 25.3 & 16.8 & 26.3 & 18.3 & 55.1 & 34.9 & 21.4 & 28.3 \\
GPT-4.1-mini     & 41.9 & 32.9 & 53.1 & 29.3 & 64.2 & 64.0 & 13.6 & 42.7 \\
GPT-5-nano     & 43.4 & 41.7 & 82.5 & 28.9 & 63.4 & 73.1 & 14.8 & 49.6 \\
\midrule
\textit{Human expert} & 82.1 & 85.7 & 82.1 & 90.9 & 92.9 & 89.4 & 95.7 & 88.4 \\
\bottomrule
\end{tabular}
}
\end{table*}

\section{Experiments and Results}
MedObvious targets a pre-diagnostic requirement, i.e., before interpretation, the model must verify that the input is coherent and safe to reason over. We therefore evaluate VLMs as potential \emph{pre-diagnostic gatekeepers} using the following clinically grounded research questions:

\noindent \textbf{RQ1 (Gatekeeping and false alarms).} Can models detect gross input violations (wrong modality or anatomy) and decide whether to proceed or abstain? Also, can models correctly determine that no anomaly is present when given normal, internally consistent inputs?

\noindent \textbf{RQ2 (Set-level consistency).} How does performance change as the candidate set grows, requiring systematic comparison across images?

\noindent \textbf{RQ3 (Clinical semantics).} Do models reliably detect clinically meaningful mismatches (e.g., anatomy/viewpoint) rather than relying on superficial cues, and do they confuse such mismatches with pathology?

\noindent \textbf{RQ4 (Grounding).} Under physically implausible or cross-modality inconsistencies, do models reject the input or rationalize it with plausible narratives?


\noindent  \textbf{RQ5 (Interface robustness).} Are sanity-check decisions consistent across binary, localization, and free-text interfaces, or strongly format-dependent?






\noindent \textbf{Evaluation Pipeline.} All models are evaluated on the full MedObvious benchmark in a \textbf{zero-shot} setting, without fine-tuning, retrieval augmentation, or few-shot exemplars. 
We evaluate three model groups:
\begin{itemize}
    \item \textbf{General open-source VLMs:} LLaVA-1.5-7B~\cite{liu2024improved}, Qwen2-VL-7B~\cite{wang2024qwen2}, \\Qwen2.5-VL-7B, Qwen3-VL-8B~\cite{bai2025qwen3}, InternVL2.5-8B~\cite{chen2024expanding}, InternVL3-8B~\cite{zhu2025internvl3}, and Pixtral-12B~\cite{agrawal2024pixtral}.
    
    \item \textbf{Medical open-source VLMs:} LLaVA-Med-7B~\cite{li2023llava}, MedGemma1.5-4B-IT~\cite{sellergren2025medgemma}, Lingshu-7B~\cite{xu2025lingshu}, and Fleming-8B~\cite{liu2025fleming}.
    
    \item \textbf{Proprietary VLMs:} Gemini-2.0-Flash, Gemini-2.5-Flash, GPT-4o, GPT-4.1-mini, GPT-4.1-nano, and GPT-5-nano.
\end{itemize}

Each instance is evaluated in five formats, each with format-specific prompts. Outputs are constrained and parsed into a closed label space (option letter, grid position, or Yes/No) to ensure consistent scoring across models.
Moreover, open-source models are evaluated with a unified inference pipeline on NVIDIA A100 (40\,GB) GPU. Proprietary models are queried via public APIs using the same prompts and output normalization.



\noindent \textbf{Evaluation Metrics.}
We report accuracy for each format, as well as \textbf{Positive} accuracy (outlier present), \textbf{Negative} accuracy (no outlier), and \textbf{Overall} accuracy. Reporting Positive and Negative separately is clinically important as a model can appear strong on anomaly-present cases, yet remain unsafe due to false alarms on normal inputs.



\noindent \textbf{Results.} We summarize results by research question. Table~\ref{tab:main_results} reports performance on the 5 evaluation modes, and Table~\ref{tab:version_evolution} reports performance on 5 tiers.

\begin{table*}[t]
\centering
\small
\caption{Per-tiers overall accuracy (\%) for different VLMs.}
\label{tab:version_evolution}

\makebox[\textwidth][c]{%

\begin{minipage}[t]{0.31\textwidth}
\centering
\setlength{\tabcolsep}{6pt}
\renewcommand{\arraystretch}{1.03}
\resizebox{\linewidth}{!}{%
\begin{tabular}{lccccc|c}
\toprule
\rowcolor{cyan!20}
\textbf{Model} & \textbf{T1} & \textbf{T2} & \textbf{T3} & \textbf{T4} & \textbf{T5} & \textbf{All} \\
\midrule
\rowcolor{cyan!10}
\multicolumn{7}{c}{\textbf{\textit{General open-source VLMs}}}\\
\midrule
LLaVA-1.5-7B    & 35.2 & 40.4 & 21.6 & 42.5 & 36.7 & 35.2\\
Qwen2-VL-7B     & 50.9 & 47.2 & 37.2 & 52.8 & 39.6 & 45.5\\
Qwen2.5-VL-7B   & \textbf{68.8} & \textbf{67.2} & \textbf{50.5} & \textbf{84.0} & 49.2 & \textbf{63.9}\\
Qwen3-VL-8B     & 47.2 & 48.9 & 34.7 & 53.4 & 32.8 & 43.4\\
InternVL2.5-8B  & 58.8 & 56.0 & 33.0 & 67.2 & \textbf{49.3} & 52.8\\
InternVL3-8B    & 50.4 & 50.4 & 37.2 & 57.1 & 33.9 & 45.8\\
Pixtral-12B     & 45.0 & 41.4 & 26.3 & 47.8 & 33.2 & 38.7\\
\bottomrule
\end{tabular}}
\end{minipage}

\begin{minipage}[t]{0.35\textwidth}
\centering
\setlength{\tabcolsep}{6pt}
\renewcommand{\arraystretch}{1.03}
\resizebox{\linewidth}{!}{%
\begin{tabular}{lccccc|c}
\toprule
\rowcolor{green!20}
\textbf{Model} & \textbf{T1} & \textbf{T2} & \textbf{T3} & \textbf{T4} & \textbf{T5} & \textbf{All} \\
\midrule
\rowcolor{green!10}
\multicolumn{7}{c}{\textit{Medical open-source VLMs}}\\
\midrule
LLaVA-Med-7B      & 32.7 & 32.5 & 28.8 & 26.8 & 25.0 & 29.1\\
Fleming-8B        & 41.8 & 43.9 & 29.1 & 39.0 & 31.4 & 37.0\\
Medgemma1.5-4B    & 59.7 & 53.7 & 37.2 & 53.7 & \textbf{53.5} & 51.5\\
Lingshu-8B        & \textbf{62.9} & \textbf{64.7} & \textbf{48.8} & \textbf{63.1} & 46.0 & \textbf{57.1}\\
\bottomrule
\end{tabular}}
\end{minipage}

\begin{minipage}[t]{0.33\textwidth}
\centering
\setlength{\tabcolsep}{6pt}
\renewcommand{\arraystretch}{1.03}
\resizebox{\linewidth}{!}{%
\begin{tabular}{lccccc|c}
\toprule
\rowcolor{orange!20}
\textbf{Model} & \textbf{T1} & \textbf{T2} & \textbf{T3} & \textbf{T4} & \textbf{T5} & \textbf{All} \\
\midrule
\rowcolor{orange!10}
\multicolumn{7}{c}{\textit{Proprietary VLMs}}\\
\midrule
Gemini-2.0-Flash   & \textbf{59.3} & \textbf{61.0} & \textbf{56.3} & \textbf{66.8} & 34.6 & 55.6\\
Gemini-2.5-Flash   & 59.0 & 62.7 & 55.2 & 62.1 & 35.0 & 54.8\\
GPT-4o             & 54.3 & 56.0 & 45.2 & 59.6 & 34.6 & 49.9\\
GPT-4.1-nano       & 29.3 & 31.8 & 20.2 & 31.8 & \textbf{37.5} & 30.1\\
GPT-4.1-mini       & 47.7 & 52.2 & 36.1 & 50.9 & 33.5 & 44.1\\
GPT-5-nano       & 52.5 & 56.6 & 46.6 & 61.2 & 33.2 & 50.0\\
\bottomrule
\end{tabular}}
\end{minipage}%

}
\end{table*}

\noindent \textit{\textbf{RQ1 (Gatekeeping and false alarms).}}
Table~\ref{tab:main_results} shows that overall accuracy is still far from a reliable pre-diagnostic gate, with large variance between models. 
The most safety-relevant signal is the \textbf{Pos(+)} and \textbf{Neg($-$)} split. Several models achieve high Pos(+) accuracy while collapsing on Neg($-$), indicating a ``\emph{always-find-something}'' bias that would be unacceptable for a gatekeeper. In contrast, a smaller subset achieves substantially higher Neg($-$), demonstrating that abstention is learnable, but not consistently present across model families. Importantly, the proprietary scale does not automatically fix this failure mode, as the negative accuracy remains modest for many such models, suggesting that normal-case calibration is a distinct problem from diagnostic fluency.

\noindent \textit{\textbf{RQ2 (Set-level consistency).}}
The tier analysis in Table~\ref{tab:version_evolution} reveals that the scaling from $2\times2$ to $3\times3$ is not a minor increase in difficulty but a \textbf{qualitative failure point}. The systematic drop in T3 (\textsc{Scaling}) suggests that many models do not reliably perform an exhaustive comparison across candidates. Instead, they appear to rely on a small number of salient cues. This matters clinically because multi-view and multi-slice studies require consistent reasoning across many related images (CT slices or multi-view fetal ultrasound), not just the detection of a single obvious frame.


\noindent \textit{\textbf{RQ3 (Clinical semantics).}}
Models often rebound on T4 (\textsc{Semantics}), which emphasizes anatomy/viewpoint mismatches. This indicates that clinically significant distribution shifts (e.g., chest vs.\ abdomen, frontal vs.\ lateral) can be easier to detect than distractor-heavy scaling, likely because they produce greater global changes in appearance. However, this ``semantic strength'' does not imply safety: a model can detect an anatomy mismatch yet still hallucinate anomalies on fully consistent grids.

\noindent \textit{\textbf{RQ4 (Grounding).}}
MedObvious includes integrity and plausibility violations to test whether models reject invalid inputs rather than rationalize them. The high false-alarm rates on negative controls, together with strong interface sensitivity (RQ6), indicate that many models do not behave as conservative verifiers and often commit to an outlier decision even when the correct response is ``none''. This is undesirable for gatekeeping, where abstention from normal or ambiguous input is often the safer behavior.

\noindent \textit{\textbf{RQ5 (Interface robustness).}}
Performance is strongly format-dependent. Many models are high on Referring MCQ (Task-C) but low on Referring Open (Task-D) (e.g., Qwen2.5-VL-7B: 75.3\% vs.\ 29.7\%; Pixtral-12B: 76.6\% vs.\ 26.3\%; Lingshu-7B: 79.5\% vs.\ 26.8\%), suggesting option selection is easier to producing grounded descriptions. The reverse asymmetry also appears as MedGemma1.5-4B-IT is strong on Detection Open (Task-B: 86.1\%) but much lower on Detection MCQ (Task-A: 23.6\%), indicating interaction between decoding and discrete choice. Overall, sanity checking is not interface-invariant; deployment may require binary gating, localization, and short explanations, so models should be evaluated for consistency across outputs rather than a single prompt style.

\noindent \textbf{Discussion.} 
MedObvious shows that fluent report-style generation does not imply reliable pre-diagnostic verification. Across models, the main failures are \textbf{false alarms} on normal grids, \textbf{degradation under scaling} when more candidates must be compared, and strong \textbf{format sensitivity} where multiple-choice can overestimate grounded ability. These failures directly limit the use of VLMs as sanity-check assistants.
The implications are even more prominent for \textit{agentic} medical AI. VLM-based agents operating in viewers such as ITK-SNAP or 3D Slicer must make decisions from multi-panel, multi-image layouts, where basic set-level consistency (modality, anatomy, orientation, integrity) is a prerequisite for safe actions. A model that hallucinates anomalies or fails under larger candidate sets can propagate errors by selecting the wrong series or acting on invalid inputs while remaining confidently fluent.
Overall, MedObvious complements existing benchmarks by isolating this prerequisite layer and motivates targeted methods to \textit{(i)} reduce false alarms via calibrated abstention and \textit{(ii)} improve systematic set-level comparison under distractor scaling.

\section{Conclusion}
\label{sec:conclusion}
We present \textbf{MedObvious}, a benchmark for pre-diagnostic visual sanity checking in medical VLMs. Across progressive tiers, formats, and negative controls, we find that current models remain unreliable gatekeepers, with frequent false alarms, scaling degradation, and strong format sensitivity, motivating pre-diagnostic triage as a prerequisite for safe clinical and agentic deployment. A limitation is the use of simplified grids. Future work should extend to full multi-series volumes and interactive viewer-based evaluation.

\bibliographystyle{splncs04}
\bibliography{references}

\end{document}